# Using Chinese Text Processing Technique for the Processing of Sanskrit Based Indian Languages: Maximum Resource Utilization and Maximum Compatibility.

*Md Maruf Hasan*
Dept. of Information System & Computer Science
National University of Singapore

*E-mail: marufhas@iscs.nus.sg*

## Abstract

*Chinese text processing systems are using Double Byte Coding , while almost all existing Sanskrit Based Indian Languages have been using Single Byte coding for text processing. Through observation, Chinese Information Processing Technique has already achieved great technical development both in east and west. In contrast, Indian Languages are being processed by computer, more or less, for word processing purpose. This paper mainly emphasizes the method of processing Indian languages from a Computational Linguistic point of view. An overall design method is illustrated in this paper.This method concentrated on maximum resource utilization and compatibility: the ultimate goal is to have a Multiplatform Multilingual System.*

## 1. Introduction

Chinese Language is being processed by computers of every platporm quite efficiently. If it can be made possible to come out with a design where other languages (especially complicated oriental languages) are implemented in the same way as Chinese language, then maximum resource utilization can be made possible and maximum compatibility can be achieved. This is the main motivation behind this research.

Sanskrit based Indian Languages, unlike English (or other simple[1] western languages), are not so easy to process in a computer because of their typical two dimensional features[2] . For this two dimensional behavior, processing of Indian languages can take advantage of Chinese language processing technique. In the rest of this paper we are going to describe the linguistic characteristics of Sanskrit based Indian languages and Chinese language, their processing techniques and the mathematical model of the proposed system as well.

## 2  Overview of Chinese Text Processing Method.

Till now, mainland China and Singapore are using GB coding [2] of Chinese ideographic character where they encoded 6763 most frequently used Chinese characters. Hong Kong and Taiwan are using BIG-5 encoding [3]. Also there are CJK ( Chinese, Japanese, Korean ) recommendation and UNICODE recommendation [4]. All these coding methods proposed DBCS (double byte coding system) to handle large volume of ideographic character.

Till now, many Chinese Language Processing sytems have been implemented. The system architecture can generally summurize as shown in Figure 1. Software Implementation can be summerized as shown in Figure 2. A brief illustration of these figures is given below. For a greater detail of Chinese system see [1].

---

[1] Simple in a sence that these languages are written as an one dimensional straight forward array of letters.

[2] 2-D features, such as, ligature, special character ordering etc are described later in this paper.

**Coding**

A double byte code is allocated, for every Chinese ideographic character. This code is called the Internal Code. Several other codes£¬like Interchange Code, QwWei Code etc. are defined for different purposes¡£

**Font**

For each Chinese ideographic character, a 16x16 bitmapped font (mainly for display) is created and saved as a binary file. 24x24, 48x48 bitmapped fonts are created in the same way and saved as a binary file (mostly for printing). Current systems are also using TrueType fonts.

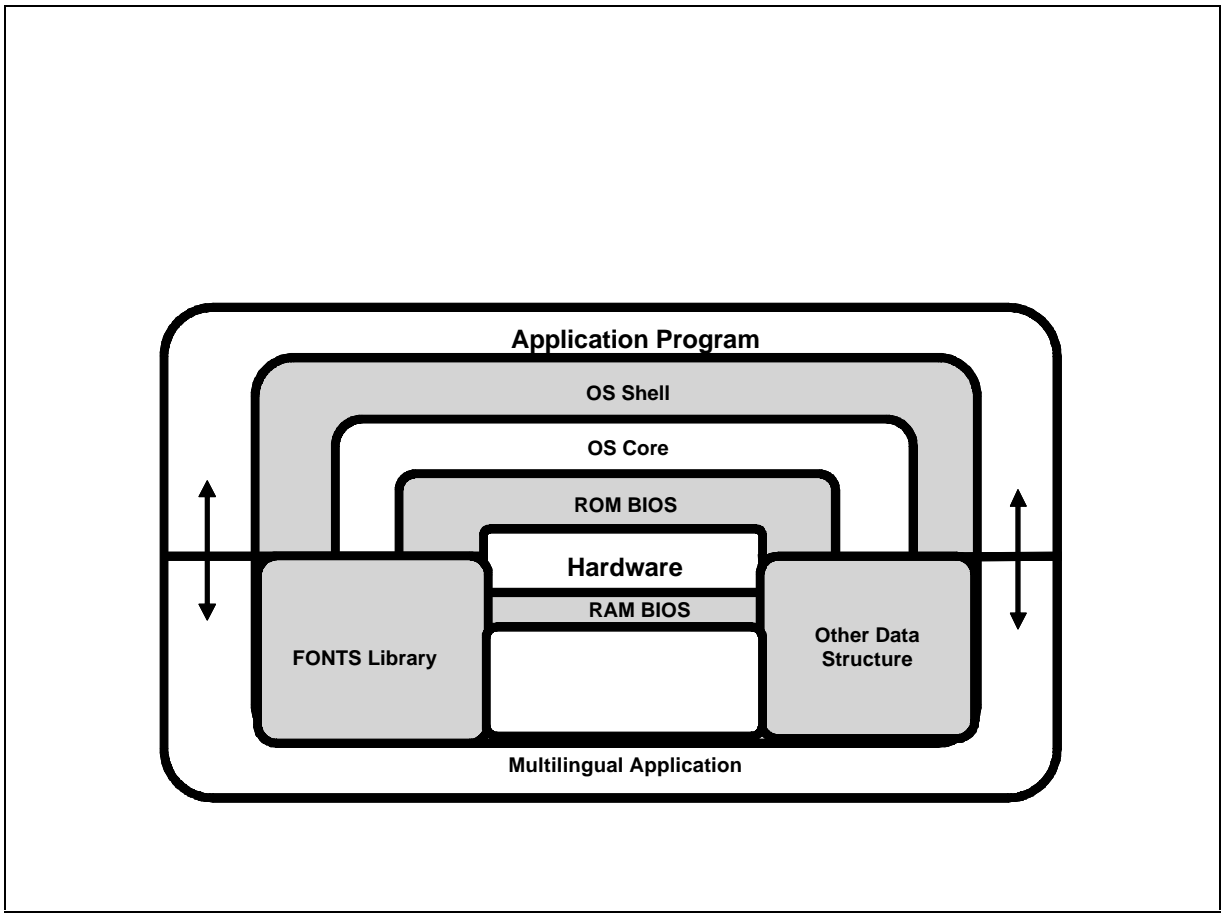

Figure 1: Multilingual Sytem Architecture

**Input Method**

Several number of input methods are currently being used with Chinese systems. Input methods are mainly a table lookup conversion process. Say PinYin-HanZi convertion input method: It provides a keyboard layout to the user using the phonetic alphabet. Then, a table of input keys and Chinese character code is being used to map the input key into a chinese character. Majority of input method is a one-to-many mapping. Prompt-line display and selection mechanism is used for disambiguation.

**Display and Print**

Display Control Module, with the help of Font Management Module generates the address of the corresponding ideographic character in the fonts library, get the respective character image from the fonts library and then pass it to the display device to display accordingly. Similar mechanism is used for printing too. Display and print are usually done in graphic mode.

**Data Communication**

Communication control module converts the Internal Code into the Interchange Code and provides other supports towards communication protocols.

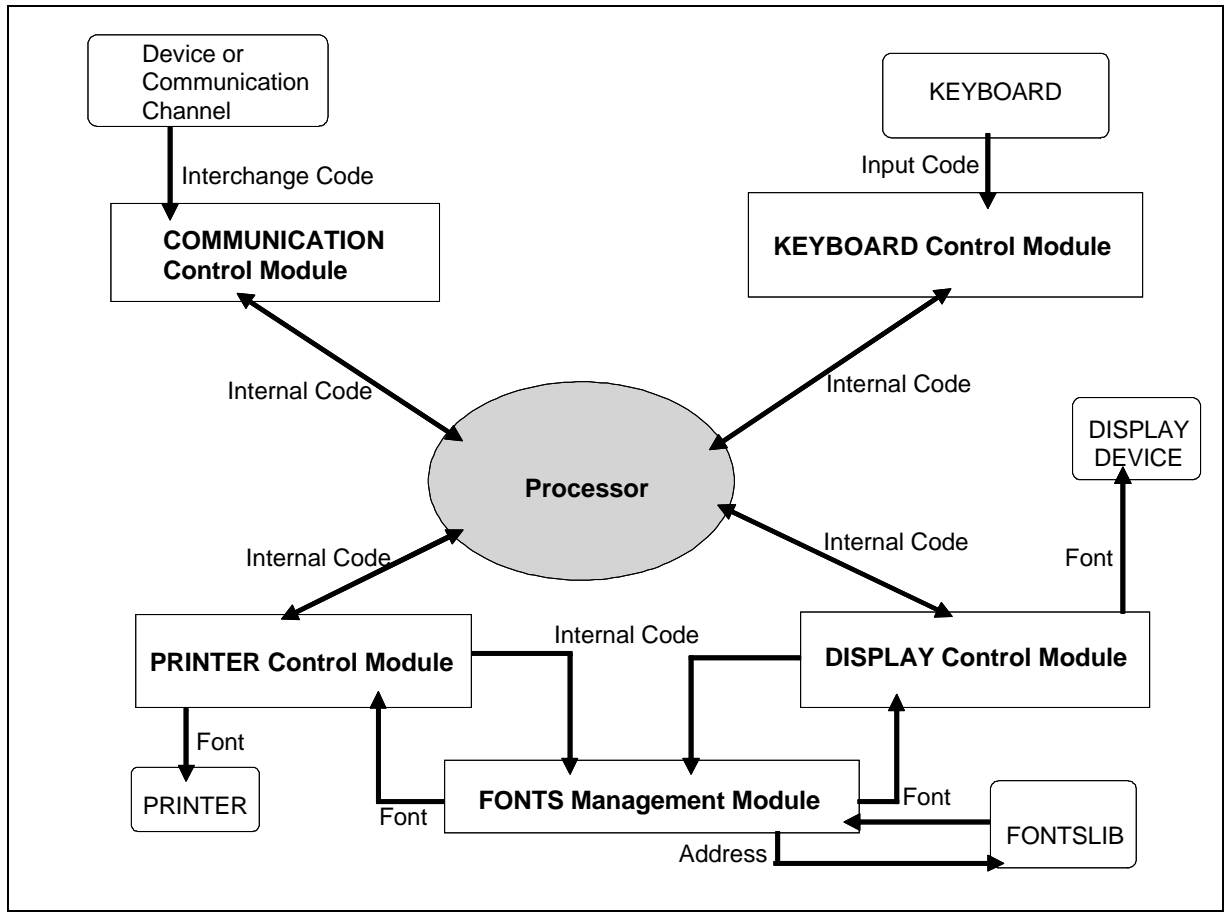

Figure 2: Software System Architecture

## 3 Sanskrit Based Indian Languages and Their Linguistic Characteristics

Sanskrit is an ancient language. Most of the Indian languages are the derivative of Sanskrit. Hindi (Indian national language), Bengali (national language of Bangladesh and Indian state of West Bengal), Nepali, Tamil, Gurumukhi, Gujarati, Oriya, Telegu, Kannada, Malayalam etc are the example of Sanskrit based languages. Hindi and Nepali are written in Devanagari script, all other languages are written in their own script. All these languages share a number of common linguistics characteristics.  These languages, unlike Chinese, possess Alphabet Set consist of  a small number of letters (more or less 50 letters, including vowels and consonants) . But again, these languages do not  follow the straight forward one dimentional array like style to form word as it is the case in English and some other western languages.

Table 1 shows the Devanagari and Bengali  Coding table recommended by UNICODE, (Currently being used in some of the existing system). The similarities of these two languages are very obvious in the table. Two dimentional characteristics are also partially visible there. Moreover, characters in the similar positions mostly share similar pronounciation too.

Although most of the Indian languages have tiny set of  Alphabet, but the glyphs of the constituent letters may take several different form in different place of occurance. Consonants and vowels, consonants and consonants may combine together  and their combined form may take quite different look. Moreover the sequence of letters not always written in straight forward order. A vowel after a consonant may appear on the top or on the bottom of that consonant and even may appear before that consonant. Some vowels may wrap its host consonant too. Formation of ligature[3]  also has this irregular characteristic.

Again,  Indian languages are also two dimensional when sorting is concern. Unlike English, consonant and vowel get different priorities in sorting. Words are sorted by taking the consonant's order as first consideration,    then    the    relevant    vowel's    order    as    a    second    consideration.    In    a

---

[3] Two or more letters combined into a single typographical form.

(Consonant)(Consonant)(Consonant)* | (Vowel)* type ligature, the second or higher order consonants also have different priority than its individual appearence or its apperance as a first consonant. A sorting example is illustrated in Table 2.

| | Devanagari | | | | | | | | Bengali | | | | | | | |
|---|---|---|---|---|---|---|---|---|---|---|---|---|---|---|---|---|
| | 090 | 091 | 092 | 093 | 094 | 095 | 096 | 097 | 098 | 099 | 09A | 09B | 09C | 09D | 09E | 09F |
| 0 | | ऐ | ठ | र | ि | ॐ | ऋ | ० | | ঐ | ঠ | র | ি | | শ্র | ব |
| 1 | ँ | ऑ | ड | ऱ | ी | ॑ | ॠ | | ঁ | | ড | | ী | | ২ | র |
| 2 | ं | ऒ | ढ | ल | ु | ॒ | ॢ | | ং | | ঢ | ল | ু | | ৩ | \ |
| 3 | ः | ओ | णा | ळ | ू | ॓ | ॣ | | ঃ | ও | ণ | | ূ | | ৩ | ট |
| 4 | | औ | त | ऴ | ृ | ॔ | । | | | ঔ | ত | | ৃ | | | / |
| 5 | ऄ | क | थ | व | ॅ | | ॥ | | অ | ক | থ | | | | | ✓ |
| 6 | आ | ख | ·द | श | ॆ | | ० | | আ | খ | দ | শ | | | ০ | ৬ |
| 7 | इ | ग | ध | ष | े | | १ | | ই | গ | ধ | ষ | ে | ৗ | ১ | l |
| 8 | ई | घ | न | स | ै | क्र | २ | | ঈ | ঘ | ন | স | ৈ | | ২ | ы |
| 9 | उ | ङ | न | ह | ् | ख़ | ३ | | উ | ঙ | | হ | | | ৩ | ० |
| A | ऊ | च | प | | ॊ | ग़ | ४ | | ঊ | চ | প | | | | ৪ | ں |
| B | ऋ | छ | फ | | ो | ज़ | ५ | | ঋ | ছ | ফ | | ো | | ৫ | |
| C | ऌ | ज | ब | | ौ | ड़ | ६ | | ঌ | জ | ব | | ৌ | ড় | ৬ | |
| D | ऍ | ज्ञ | भ | ऽ | ् | ढ़ | ७ | | | ঝ | ভ | | ্ | ঢ় | ৭ | |
| E | ऐ | ञ | म | ा | | फ़ | ८ | | | ঞ | ম | া | | | ৮ | |
| F | ए | ट | य | ि | | य़ | ९ | | এ | ট | য | ি | | | ষ | ৯ |

.

Table 1 : Devanagari and Bengali Codepage

**Source**: The Unicode Standard.

Considering all these typical characteristics, a mathematical model of Indian language processing is built to avoid the complexity of computer processing of Indian languages and to take the advantage of Chinese text processing technology. See Section 5.1 below.

## 4 Overview of Existing Indian Language Processing Method

Existing Indian language processing systems are mainly serving the purpose of word processing. Most of the systems are using the glyphs of the letters and a few of their variations and build up their font set of above hundred glyphs. This character set then encoded in a Single Byte space (most of the systems are making use of ASCII extention code space). A keyboard layout, based on the glyphs, is provided to the user to input symbols. [Note that these systems failed to maintain typical written form of ligatures due to a small number of available glyphs.]

Table 2: Example of 2-D Sorting

Some of the systems are making use of Text Rendering Techniques (See Figure 3) to minimise distortion of script and to maintain 2-D arrangement of letters.

Obvious enough that implementation of sorting algorithm become very complicated in these system. Most of the existing systems simply avoided sorting facilities.

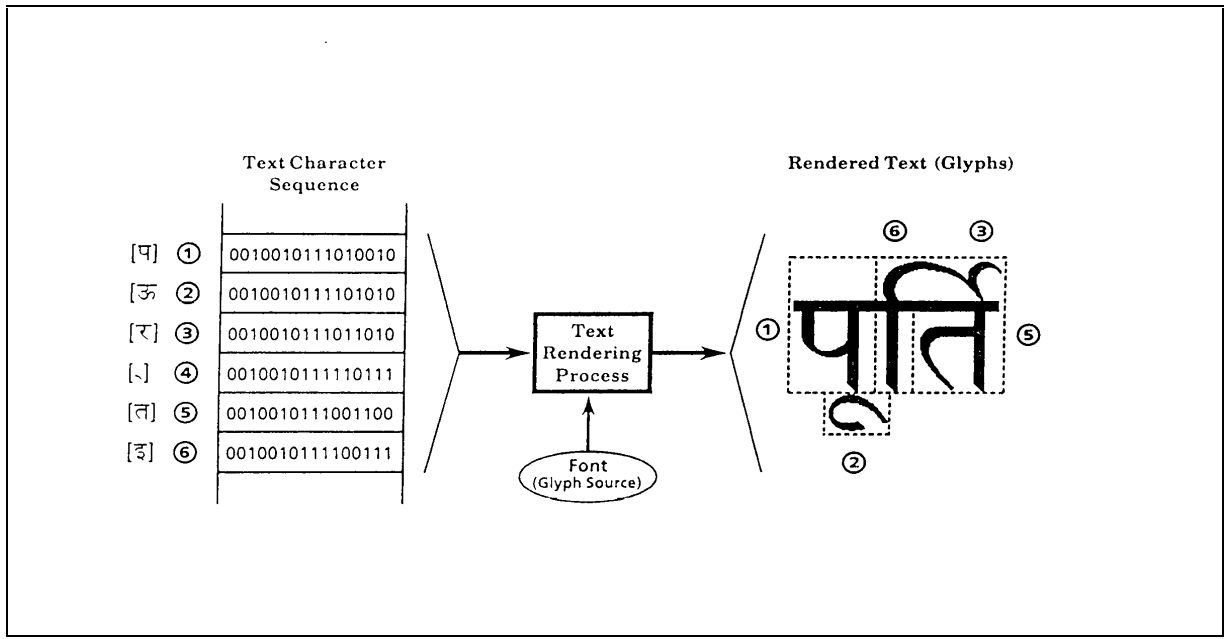

Figure 3: Text Rendering of Devanagari Text

**Source**: The Unicode Standard.

The existing system might be described easily by comparing it with the folowing approach of Chinese text processing.
- Encode and design glyphs of Redicals like 口 艹 女 力 田 .............
- Use text rendering to create characters like 口 如 男 茹 ..................
- Process Chinese text.
   It's somehow Possible!

People can easily imagine the complexity of such an approach, specially the complexity of
- Text Rendering Algorithm
- Sorting Algorithm   etc.

Moreover, the implemented systems are designed on the application layer (see Figure 1), which is totally unsupportable in computational linguistic point of view. As a result, these systems fail to provide database, communication and other facilities. An OS level design is desirable to provide these facilities.

## 5  Detail of New System Design

We used Bengali language as an example Sanskrit based language and described design and implement issues of our system. The system is more like a CCDOS[4] (Chinese Character Disk Operating System) system [5]. CCDOS encoded a total of 6763 Chinese characters. According to GB2312-80 [2], the first level Chinese character consist of 3755 characters which cover 95% character of modern Chinese text. The next level consists of another 3008 Chinese characters which cover another 5% of modern Chinese text. As an experiment, we just replaced the Second Level character fonts from CCLIB (CCDOS fonts library) with Bengali character fonts. Then we designed an External Input Module[5] where we used a new convertion table and search algorithm to manipulate bilingual input. Output (especially display and print) mechanism need no modification. This system successfully provided multilingual (Bengali, Chinese and English ) environment in a PC. Other Localized Chinese software like Chinese Dbase, Chinese Word Star also supported the multilingual environment.

Chinese hardware like HanKa (Chinese Card) can also be used with our system. The only necessity is the replacement of Chinese font containing EPROM chip with a bilingual font containing EPROM chip. The maximum compatibility and maximum resourse utilization come here.

### 5.1   Mathematical Model of Double Byte Processing of Indian Languages

Here we are giving the mathematical model of our Bengali text processing system. Other Sanskrit based Indian languages can also be modelled in the same way and can share same processing techniques. In this model, we treat letters in Bengali Alphabet Set as redicals in Chinese language. Then we collect all possible ligatures through linguisttic analysis. We treat these ligatures (along with the independent vowels and consonants) in the same way as the Chinese characters are treated in a chinese text processing system, finally we process them in the same manner as the Chinese characters are processed

**Basic Definitions**

**Definition 1.** A consonant in Bengali will be represented as $c_i$ , i=1....39. (There are about 39 commonly used consonants in Bengali ).
   **Constant Set, C = {$c_i$ }**

**Definition 2.** An Independent Vowel (Dependent Vowels are the symbolic variations of independent vowels usually appear with consonants/ligatures ) will be represented as $v_j$ , j=1.....11   (There are about 11 commonly used vowels in Bengali).
   **Vowel Set, V = {$v_j$}**

**Definition 3.** The combination of one consonant and one vowel; one consonant and one diacritical mark; two ore more consonants with (or without vowel / diacritical mark) will be called ligature and represented as $l_k$,

---

[4] CCDOS is a software copyrighted to 北京电子工业部。
[5] CCDOS facilitated External Input Module Interface

k=1.....2,500. (We analysed bengali language and found that there are about 2,500 commonly used ligature in Bengali)
   **Ligature Set, L = {$l_k$}**

**Definition 4.** Including Bengali digits, monetary symbol etc there are about 20 symbols in Bengali, which will be represented as $S_l$, l=1.....20.
   **Symbol Set, S = {$s_l$}**

**Definition 5.** Word Constituent Unit ($u_m$) is defined in the following way:
$u_m \rightarrow c_i | v_j | l_k$, generally m<=3000
   **Word Constituent Unit Set, U = {$u_m$}**

**Definition 6.** Word will be represented as $w_n$,
n=1.....α virtually.
$w_n = (u_m)(u_m)^*$
   **Word Set, W = {$w_m$}**

**Definition 7.** We denote B as a set of Bengali Characters as below.
   Bengali Character Set,
   **B = { $b_i$ | $b_i \in U$ or $b_i \in S$ }**

### Mathematical Model of Double Byte Bengali Code

**Definition 8.** Each element in B is assigned a unique double byte (16 Bit) Internal Code, $i_i$.
There exists a function σ, so that
   $i_i = \sigma(b_i)$ and
   $b_j = \sigma^{-1}(i_j)$.
   Internal Code Set,
   **I = { $i_i$ | |$i_i$| = 16 bits. }**
Also if $b_i$ appears before $b_j$ in dictionary, then corresponding $i_i$ and $i_j$ will satisfy the following relation.
   $i_i < i_j$

### Mathematical Model of Bengali Character Input

Different Input methods can be designed for this Bengali system. We designed an input method, IAYS (Input As You Spell), where the user will be provided a keyboard layout (dynamic[6]) of the vowel consonant and symbols. To input Bengali, user will type in the sequence of consonants and vowels as s/he spells. For some cases, a selection option will appear in the prompt line for disambiguation. Apparently it may seem that selection keys are an extra overhead, but actually input method's performance can be highly optimized using Associative Rules. Again word input, phrase input and sentence input methods can be designed to obtain amazing input rate , key stroke ratio.
   This input method uses the table lookup mechanism , where we have a table of Input Code (Spelling Attribute) and Internal Code.

**Definition 9.** Spelling Attribute Set,
   **A = {$a_i$ | $a_i \in C, a_i \in V$}**

**Definition 10.** Input Method is an one-to-many mapping ρ, where
   $b_i = \rho(a_j)$.

### Mathematical Model of Bengali Character Output

Output process involve with another one-to-one mapping, θ, which maps the unique Internal Code Set into Fonts Attribute Set. Display Control Module is the same with Chinese system.

---

[6] We call it Dynamic; because when a user will use this system, he will be prompted to configure his own keyboard. He can also modify the keyboard layout anytime by running an utility program.

**Definition 11**. Fonts Attribute Set,
F = { $f_i$ | $f_i$ is a binary(0,1) sequence of
16*16, 24*24 bits}

**Definition 12**. Output of a Bengali Character is a one-to-one mapping, θ, where
$f_i = θ(i_j)$.

## 6  Conclusion

This is the very first implementation of Indian language in the Chinese Text Processing skeleton. Although the ligature analysis, font design, etc are not so efficient but the approach we made here is very prospective in a sense that it is the OS level implementation of Indian language. Again, on the strong technical foundation of Chinese processing, such Indian language processing systems will inherit a lot of software and hardware resources just after its birth. Moreover such approach will decrease difficulty to maintain compatibity between different text processing systems. The ultimate goal is a Global Text Proccessing System in a common technical frame.

It is very obvious that the mathematical model of the Chinese system is very similar to that of our Bengali system. Thus it becomes very easy to port our Bengali system in other platforms where Chinese language has already been implemented.

**Acknowledgements**

I like to thank Mr Kawser and Mr Ashraful to help me in linguistic analysis and font design; Prof Mao Yu Heng , Department of Automation, Tsinghua University, China, to provide Font Design Tools and  Mrs Dai Mei Er, Assosiate professor, Department of Computer Science, Tsinghua University, for her kind supervision.

I am also indebted to Dr Lua Kim Teng, my present supervisor, who always reminds me to do something for the development of my mother tongue. Thanks for all the users who kindly used the system and obliged me with their valuable comments for further development.